\documentclass[preprint,5p,times,twocolumn]{elsarticle}

\usepackage{amssymb}

\usepackage{graphicx}
\usepackage{amsmath}
\usepackage{algpseudocode}
\usepackage{algorithm}
\usepackage{slashbox}
\usepackage{mathtools}
\usepackage{multirow}

\usepackage{tabu}
\usepackage{colortbl}
\usepackage{fancyhdr}
\usepackage{hyperref}

\usepackage{color}

\newcommand{\newtext}[1]{\textcolor{black}{#1}}
\newcommand{\newtextRed}[1]{\textcolor{black}{#1}}

\journal{Computer Vision and Image Understanding}

\begin{document}

\begin{frontmatter}



\title{Context-based Object Viewpoint Estimation:\\ A 2D Relational Approach}


\author[ESAT]{Jos{\'e} Oramas M.}
\author[CS]{Luc De Raedt}
\author[ESAT]{Tinne Tuytelaars}



\address[ESAT]{KU Leuven, ESAT-PSI, IMEC\\
Kasteelpark Arenberg 10 - bus 2441\\
B-3001 Heverlee, Belgium}

\address[CS]{KU Leuven, CS-DTAI\\
Celestijnenlaan 200A\\
B-3001 Heverlee, Belgium}

\begin{abstract}
   The task of object viewpoint estimation has been a challenge since the early days of computer vision.
   To estimate the viewpoint (or pose) of an object, people have mostly looked at object intrinsic features, such as shape or appearance. 
   Surprisingly, informative features provided by other, extrinsic elements in the scene, have so far mostly been ignored.
   At the same time, contextual cues have been proven to be of great benefit for related tasks such as object detection or action recognition.
   In this paper, we explore how information from other objects in the scene can be exploited
   for viewpoint estimation. In particular, we look at object configurations by following a relational neighbor-based approach for reasoning about object relations.
   We show that, starting from noisy object detections and viewpoint estimates, exploiting the estimated viewpoint and location of other objects in the scene can lead to improved object viewpoint predictions.
   Experiments on the KITTI dataset demonstrate that object configurations can indeed be used as a complementary cue to appearance-based viewpoint estimation.  
   Our analysis reveals that the proposed context-based method can improve object viewpoint estimation by 
   reducing specific types of viewpoint estimation errors commonly made by methods that only consider local information. Moreover, considering contextual information produces superior performance in scenes where a high number of object instances occur. 
   Finally, our results suggest that, following a cautious relational neighbor formulation 
   brings improvements over its aggressive counterpart for the task of object viewpoint estimation.   
\end{abstract}


\end{frontmatter}

\thispagestyle{fancy}



\section{Introduction}
During the last decade, contextual information has proven beneficial for 
vision tasks such as image segmentation and object detection. 
For the task of object detection, there is a significant amount of work, e.g. 
\cite{LAntanas:Neurocomputing,SUN09ChoiLTW10,cinbisS12,DesaiRF11_ijcv,dkollerECCV2008,perkoCVIU2010,songCHHY11}, in which pairwise relations between object hypotheses are exploited to re-rank the initial predictions given by the object detector. 
Following a different direction, a more recent group of works 
\cite{AlexeNIPS12,OramasCVIU16,QuinnRM16} has focused on exploiting 
contextual information to iteratively generate object proposals during 
test time and improve object detection.
Likewise, for image 
segmentation \cite{cgalleguillosCVPR2010,JainGD10_ECCV,tmalisiewiczNIPS2009}, 
context is considered by analyzing appearance and spatial co-occurrence of 
neighboring segments and is used to enforce spatial consistency. 
However, despite the demonstrated benefits for the already mentioned tasks, contextual 
information has been mostly ignored for the task of object viewpoint or pose 
estimation. Only recently, \cite{joramas:ICCV13}, \cite{xiang_3drr13} and 
\cite{ZiaCVPR14} took initial steps towards exploiting contextual information 
for predicting the viewpoint/pose of a group of objects.

\begin{figure}[t!]
\centering
\includegraphics[width=0.48\textwidth]{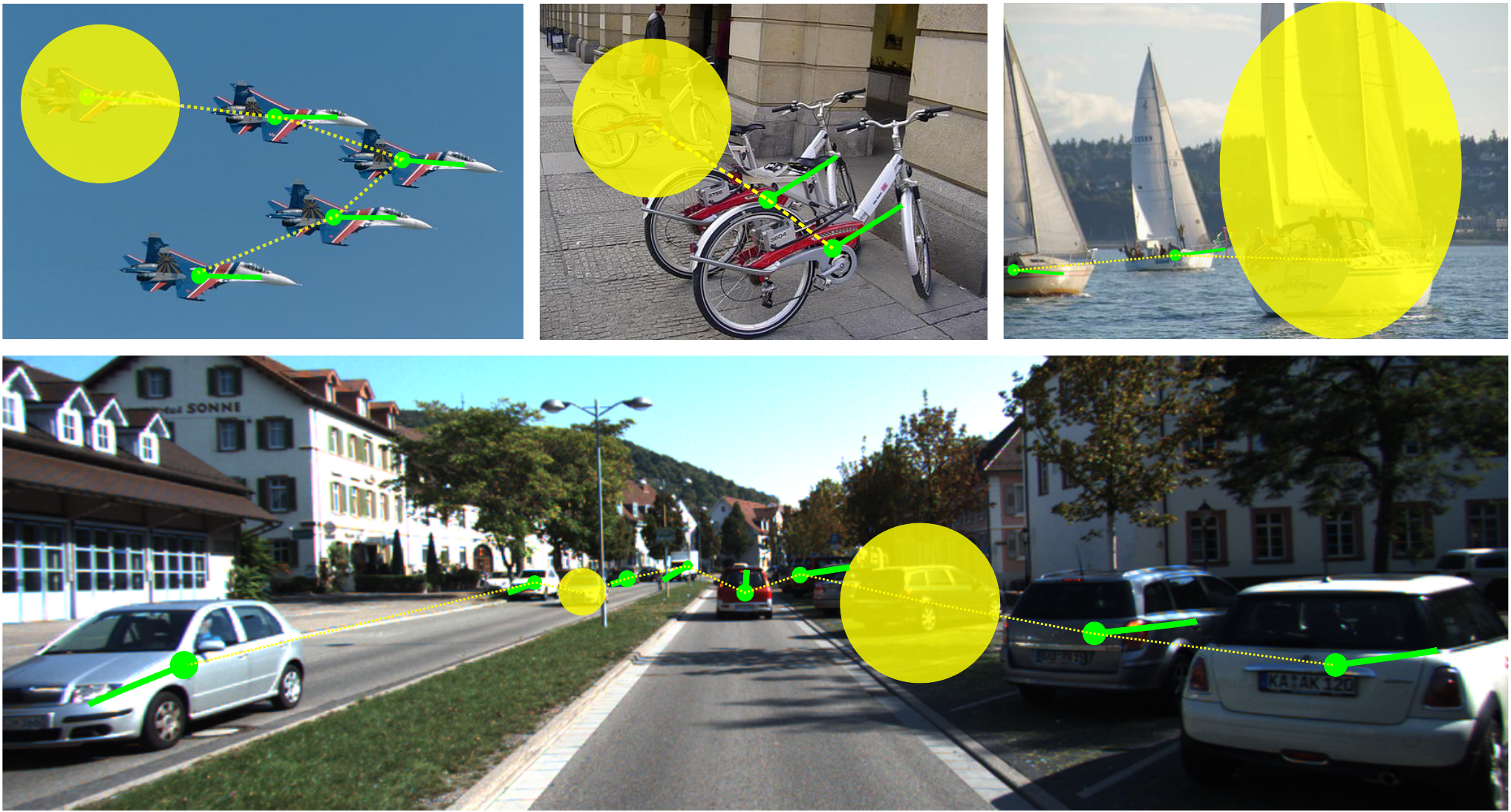}

\caption{ The natural or ``desired'' configurations in which objects 
occur in the world often provide strong cues of their viewpoint. For instance, 
it is not difficult to guess the viewpoint of the objects behind the yellow 
regions by only looking at the other objects in the scene.} 
\label{fig:problemStatement}
\end{figure}

\begin{figure}
\centering
\includegraphics[width=0.3\textwidth]{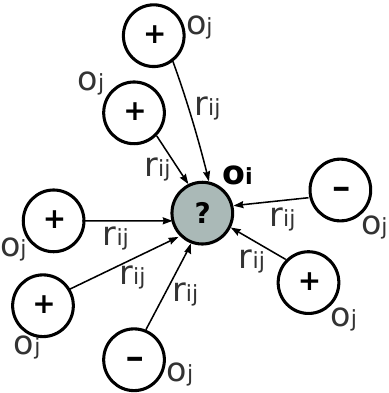}

\caption{Collective Classification. We address the classification of 
a particular object $o_i$ (in gray) based on the relations $r_{ij}$ with its 
neighboring objects $o_j$.} 
\label{fig:aggressiveInference}
\end{figure}

Here, we follow the line of our earlier work \cite{joramas:ICCV13} and 
exploit pairwise relations between objects as a source of contextual 
information for estimating the viewpoint of each of the objects.
Let us clarify the intuition behind this work with the following example:
imagine you are given the task of predicting the viewpoint of the objects below the 
yellow regions in Figure~\ref{fig:problemStatement}.
Even when there is no access to intrinsic features of the objects such as 
color or texture, the overall configuration of surrounding objects provides 
a strong cue to predict their viewpoint. This can be considered a Collective 
Classification problem \cite{SenNBG10,Senaimag08}, in which the 
class (viewpoint) of one  object influences that of another (see Figure 
\ref{fig:aggressiveInference}).
Collective classification is a popular problem in machine learning and 
data mining, in which the data takes the form of a graph and the task 
is to predict the classes of the nodes in the graph while using the 
structure of the network and a few example classifications of nodes. 
See Algorithm \ref{alg:collectiveClassification} for a brief 
description of Collective Classification \cite{SenNBG10}.

A common practice \cite{Macskassy03asimple,MacskassyP07,Wang2013MRN} 
in Collective Classification \cite{SenNBG10} when reasoning about 
relations between objects is that, during inference, the neighboring 
object hypotheses are considered without taking into account the 
certainty of their prediction. As a result, \textit{all} the neighbors 
participate in and contribute equally to the classification of 
each object.
\mbox{Following} the literature 
\cite{McDowellCautiousInfGA07,McDowellCautiousGA09} on Collective
Classification, instead, we propose an iterative scheme 
where we first classify the viewpoints of objects with the most 
certain relational information, and then use these to bootstrap 
the predictions of the other objects. This is useful in collective 
classification tasks, like object detection or object viewpoint 
estimation, where multiple possibly related objects all need 
to be classified (see Figure \ref{fig:problemStatement}). 
Following the terminology of \cite{McDowellCautiousInfGA07}, we 
refer to these two inference variants as ``aggressive'' inference, 
where all the neighboring objects are considered as sources of 
contextual information, and ``cautious'' inference, where we 
iteratively select the objects with highest certainty as source 
of contextual information. In this paper, we empirically evaluate 
the added value of aggressive vs. cautious inference for the 
task of object viewpoint estimation.

\begin{algorithm}
\caption{\label{alg:collectiveClassification} Collective Classification~\cite{SenNBG10}}
\textbf{\footnotesize Given}{\footnotesize \par}
\begin{itemize}
\item {\footnotesize A set of interrelated nodes $o_i$ connected by links $r_{ij}$.}{\footnotesize \par}
\end{itemize}

\textbf{\footnotesize Steps}{\footnotesize \par}
\begin{enumerate}
\item {\footnotesize \textbf{Local Classification:} classify each of the nodes $o_i$ using the non-relational (local) model. This model focuses purely on attributes of the nodes.}{\footnotesize \par}
\item {\footnotesize \textbf{Relational Classification:} classify each of the nodes $o_i$ using the relational classifier which takes into account neighboring nodes $o_j \in N_i$ connected via links $r_{ij}$.}{\footnotesize \par}
\item {\footnotesize \textbf{Collective Inference:} re-classify the nodes $o_i$ together, taking into account the classification output obtained in steps 1 and 2, and possibly iterate.}{\footnotesize \par}			  			  
\end{enumerate}
\end{algorithm}

\newtext{
This paper complements our earlier work \cite{joramas:ICCV13} in two ways: first,
by providing a more elaborate theoretical treatment of the method, and 
second, by providing a wider experimental validation, including an evaluation 
with a larger set of detectors, and a deeper analysis of success and failure cases.
Moreover, in order to generalize to images where extracting 3D information might be 
too complex, in the present paper we focus on 2D spatial relations instead of 
relations in the 3D space.
}

This paper is organized as follows: Section~\ref{sec:relatedWork} presents related
work. 
Section~\ref{sec:proposedMethod} shows how we define and learn relations 
between objects in the scene
and how we combine the contextual response provided by the related objects 
with the evidence from local detectors. Implementation details are presented in 
Section~\ref{sec:implementationDetails}. Then, Section~\ref{sec:evaluation} 
describes our evaluation protocol 
and the obtained results and discussions. Limitations and directions 
for future work are presented in Section~\ref{sec:limitationsFutureWors}.
Finally, we draw conclusions in Section~\ref{sec:conclusions}.

\section{Related Work}
\label{sec:relatedWork}

The task of object viewpoint estimation can be analyzed from different perspectives. 
In this work we focus on four aspects that will help position our work: 
the cues used for viewpoint estimation, sources of contextual information, how inference 
between objects is performed and by comparing our work w.r.t. holistic scene understanding.

\subsection{Cues for object viewpoint estimation}
Several object viewpoint estimation methods have been proposed in the literature.
Most of these rely on intrinsic characteristics of the object category such as color,
texture or gradient patterns.
In the traditional processing pipeline for object viewpoint estimation, first, 
candidate regions to host object instances are proposed. Secondly, 
an appearance descriptor is computed in the area of each candidate 
region. Finally, based on a pre-trained model, each descriptor is 
classified as one of the possible viewpoints the object may take. 
Following this pipeline, methods have evolved from modeling the 
appearance of 2D views that the objects may take under different viewpoints 
(e.g. \cite{GGABS11,rlopez_2011,xiang_wacv14}) to reasoning 
about geometric configurations of parts of the objects in the 3D 
space \cite{Hoiem:pose3D:CVPR07,LS10,Pepik,savarese:iccv2007}. 
\newtext{
Recently, several works \cite{ghodratiBMVC14,JuranekHZICCV15,TulsianiCVPR15,Su_2015_ICCV} 
have demonstrated the benefits of using learning-based representations 
to model object appearance and perform viewpoint estimation. 
In \cite{ghodratiBMVC14}, it was shown that the activations 
of the last hidden layer of a pretrained convolutional neural network (CNN) can 
effectively serve as features to describe object viewpoints. 
Moreover, these learned features 
outperform methods that rely on traditional 
2D features and features derived from 3D models. In \cite{TulsianiCVPR15}, 
two CNN architectures are proposed, first, to coarsely estimate the 
viewpoint of the object, and second, to refine the predicted viewpoint 
by focusing on keypoints. In \cite{JuranekHZICCV15} a compact model is proposed 
where image features extracted by the detector are shared with the 
viewpoint estimator with the objective of achieving real-time 
performance. Finally, \cite{Su_2015_ICCV}  
generates synthetic images from 3D object collections to render a 
high volume of images with high variation which can be 
exploited by a CNN. 
In our earlier work \cite{joramas:ICCV13}, we explored an ``allocentric'' approach
in which the 3D pose of an object influences that of another.
In that formulation, pairwise relations between objects in the 3D
space were considered as sources of contextual information. 
To this end, a pair of local and contextual responses 
were computed for each object in the scene.  
On the local side, the output of a viewpoint-aware object detector, namely
\cite{GeigerNIPS11} and \cite{rlopez_2011}, was used. On the contextual side, 
the weighted sum of the belief of each of the contextual objects 
was computed following a weighted-vote relational classifier 
\cite{MacskassyP07}. Then, these responses were combined to produce 
a final pose prediction. Following this approach it was shown that 
considering contextual cues brings improvements to the estimation 
of object poses.
Parallel to this, \cite{xiang_3drr13} proposed a method that reasons about
intrinsic features from the objects such as the appearance of 
patches taken from the object, and contextual cues such as 2D
occlusion of other objects, to estimate the location and viewpoint of the 
objects in the scene. Based on these components, \cite{xiang_3drr13} 
proposed a spatial layout model that enforced scene consistency 
based on the 3D aspectlets of individual objects with 
object-object consistency in the form of occlusion reasoning.
Similarly, \cite{ZiaCVPR14} used a fine detail shape representation 
based on {\small CAD} models. This representation improved model-object 
matching in the scene and, as consequence, better reasoning about object
support on the ground-plane and mutual occlusion between objects. 
}
In this work we follow this line of work. 
We further explore the allocentric approach from  
\cite{joramas:ICCV13} where the object viewpoint estimation 
task is formulated as a collective classification problem. 
However, different from \cite{joramas:ICCV13}, 
we focus on the prediction of the viewpoint, i.e. the projected 
orientation of the object that is observed by the camera, rather 
than the 3D pose (azimuth angle) of objects in the 3D scene.
Different from \cite{ZiaCVPR14} and \cite{Su_2015_ICCV} we will limit 
our training procedure to focus on manually annotated images 
and not look into the use of synthetic data.
Furthermore, our method operates in still images and does not 
require image sequences as in some {\small SfM}-based approaches 
\cite{Bao_CVPR2012_SSFM,Bao_CVPR2011_SSFM}.

\begin{figure*}[t]
\centering
\includegraphics[width=1\textwidth]{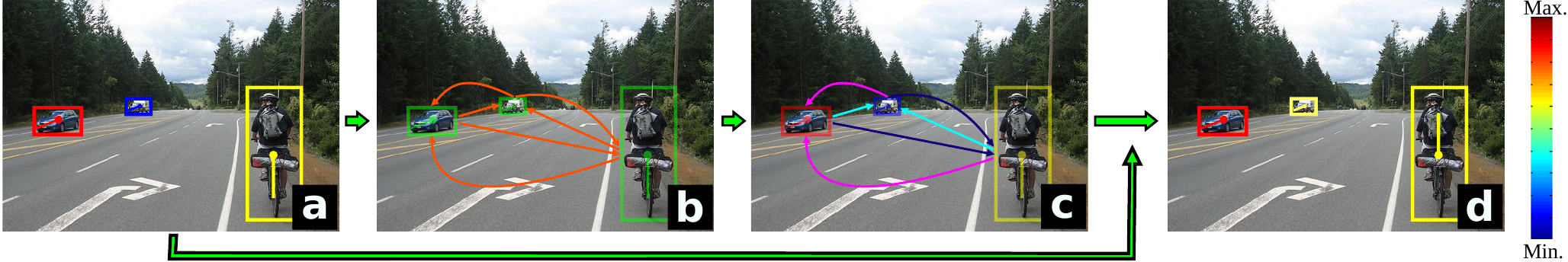}

\caption{Algorithm Pipeline: a) viewpoint-aware object detection (score encoded in jet scale), 
b) pairwise relations definition, c) object contextual scoring via \textit{wvRN} classifier \cite{MacskassyP07},
and d) combination of local and contextual responses. During the contextual scoring
step (c), the relations that are used to compute the contextual score of each object 
are grouped by color.} 
\label{fig:algorithmPipeline}
\end{figure*}

\subsection{Sources of contextual information}
Scene elements have 
been defined in several ways; \cite{Forsyth96findingpicturesTS} proposes 
that some objects have a defined shape or appearance while others can be 
characterized by their color or texture. Following these definitions 
\cite{Forsyth96findingpicturesTS} divided such elements as \textit{Things} 
and \textit{Stuff}, respectively. Both types of elements have been used in 
the past as sources of contextual information. However, in this paper we will
focus on \textit{Things}. Using this type of element, 
\cite{DesaiRF11_ijcv} defined a template on top of the bounding boxes covering
the objects. Then, these templates were used to extract discrete spatial 
relations, such as \textit{on-top, next-to, below, near, far}, between object 
hypotheses. In that work, using other objects as context proved to be helpful
to possibly identify and degrade false hypotheses bringing improvements on
object detection performance. 
Following this trend, Felzenszwalb et al. \cite{felzenszwalbTPAMI2009}, 
Perko \& Leonardis \cite{perkoCVIU2010} and Choi et al. \cite{SUN09ChoiLTW10} 
addressed a similar problem with the difference that they defined continuous spatial 
relations instead of discrete ones. These continuous relations were extracted 
by estimating the difference between the centers of object bounding boxes. Finally, 
the learned relations were used to filter out the out-of-context objects. 
\cite{songCHHY11} suggested a joint detection-classification scheme to 
identify ambiguously scored hypotheses and used an adaptive method to 
exploit context information on these hypotheses. 
In \cite{cinbisS12} relations between objects extended
the traditional approach of exploiting object co-occurrences and considered 
additional features such as relative scales, bounding box overlap ratio and scores. 
Furthermore through a set-based formulation this method has been shown able to reason 
about object spatial configurations that go beyond pairwise interactions.
Similar to these works, we exploit relations between \textit{Things}  
as sources of contextual information. However, we will focus on the 
task of estimating the viewpoint of each object. 

\subsection{How inference between objects is performed}
From the perspective of the Collective Classification literature \cite{SenNBG10}, 
specifically on the inference side, our method is inspired by 
\textit{Cautious Inference} \cite{McDowellCautiousGA09,nevilleJensenIterativeInf}. 
This is a type of inference that seeks to identify and exploit 
the more certain relational information.
Such a cautious approach was used in \cite{JainGD10_ECCV} for 
labeling object superpixels. In \cite{JainGD10_ECCV}, 
discriminative relations were mined between object regions and 
discriminative attributes were discovered per relation.
In \cite{joramas:WACV14}, relations between objects were considered 
to improve object detection by penalizing out-of-context object hypotheses. 
During inference, the object hypotheses with highest certainty
are classified first and then used to bootstrap the other objects.
Based on their experiments, it was concluded that all cases following
this cautious iterative approach resulted in better object detection
performance than when considering \textit{all} the neighboring
objects at once.
In this work, we start from the observations of \cite{joramas:WACV14},
and explore a cautious counterpart of \cite{joramas:ICCV13} 
with the objective of verifying whether the observations made for object
detection also hold for the task of object viewpoint estimation.
To this end, we first estimate the viewpoint of the objects with higher 
certainty and then use these objects to predict the viewpoints of the other
ones.

\newtext{
To some extent, the proposed method bears some resemblance to the 
message-passing algorithms \cite{Yedidia2011} commonly used for inference 
in graphical models. 
In the same fashion as the message-passing algorithms, we start from input 
elements for which information is available, in this case the objects that  
define the nodes in the graph. Then, we iteratively select a target node, 
in our case each of the object hypotheses to be re-estimated, and each of 
the neighboring nodes (objects) casts a vote (or sends a message) indicating 
the level of agreement they have with the target node taking a specific state.
Furthermore, similar to message-passing algorithms, for the case of loopy 
graphs the proposed method aims at providing an approximate solution to 
the problem at a relative low computation time. 
Different from message-passing algorithms, which operate over a graph defined 
over two types of nodes (variable nodes and check nodes), our method only 
considers variable nodes.
}

\subsection{Towards holistic scene understanding}
In the literature there is a group of works \cite{FidlerCVPR2012,HoiemEHCVPR08,kim2012eccvws,LiTPAMI2011232,SunBSIJCV12}
that study the problem of holistic scene understanding. 
This problem consists of jointly reasoning about regions, location, 
category and spatial extent of objects in the image, 
as well as the scene type. The main idea behind
holistic scene understanding is that all these problems, 
traditionally addressed in isolation, complement each other 
and this complementarity can assist to improve each individual 
task. Related works that follow these characteristics are
\cite{xiang_3drr13} and \cite{ZiaCVPR14}. These works propose
methods that, as part of their pipeline, reason about object viewpoints/poses. 
In \cite{xiang_3drr13}, a Spatial Layout Model  that jointly
reasons about inter object occlusions, rough 3D shape of
the objects and scene ground-plane is proposed. In addition, 2D 
regions partially covering the objects are grouped producing
``3D aspectlets''. The method proposed in \cite{ZiaCVPR14} 
reasons about fine-detailed 3D shape of the objects while
reasoning about object occlusions and scene ground-plane
contact. Different from the approaches addressing the 
more general problem of holistic scene understanding,
our approach is more specific in the sense that it focuses
purely on reasoning about relations between objects, i.e. \textit{Things}
entities. Particularly, in comparison with \cite{xiang_3drr13}
and \cite{ZiaCVPR14} which relate objects as means to model
occlusions, in our work relations between objects are aimed
at modeling usual configurations in which the objects of interest
occur.
Our method 
represents a more semantic relational layer between 
objects that could be integrated in current 
methods aiming at holistic scene understanding.

\section{Proposed Method}
\label{sec:proposedMethod}
This work is based on our previous work~\cite{joramas:ICCV13}.
While in \cite{joramas:ICCV13} we reasoned 
about locations and poses of objects in the 3D scene, here we shift 
the feature extraction and reasoning to the 2D image space.
We assume that some features of an object, in this case its 2D location 
and viewpoint, are not only influenced by the object itself but somehow 
driven by other entities in the scene. This idea is inspired by the concept of ``Allocentrism'',
a term in Psychology used to define entities that 
tend to be interdependent, defining themselves in terms of
the group they are part of, and behaving according to the
norms of the group \cite{Hulbert:2001:allocentrism,Triandis_Suh_2002:allocentrism}. 
Allocentric entities appear to see themselves as an extension 
of their group. Based on this description, our method takes 
into account the group consistency of each entity relative to the 
group defined by the other entities in the scene.

The proposed method consists of four steps (see Algorithm~\ref{alg:proposedMethod}):
First, given an image, we run an off-the-shelf viewpoint-aware object 
detector to collect a set of object hypotheses with class label and 
predicted discrete viewpoint
(Figure~\ref{fig:algorithmPipeline}(a)). Then, we define
pairwise relations between all the object hypotheses 
(Figure~\ref{fig:algorithmPipeline}(b)). Third,
for each of the object hypotheses, we estimate its contextual
response using as source of contextual information the other
object hypotheses (see Figure~\ref{fig:algorithmPipeline}(c)). 
Finally, we combine the local response, provided by the viewpoint-aware 
object detector, with the contextual response to obtain the 
final viewpoint estimate (Figure~\ref{fig:algorithmPipeline}(d)).
Now we will present a more detailed description of the proposed
method.

\begin{algorithm}
\caption{\label{alg:proposedMethod} Proposed Method }
\textbf{\footnotesize Given}{\footnotesize \par}
\begin{itemize}
\item {\footnotesize viewpoint-aware object detector.}{\footnotesize \par}
\item {\footnotesize Image $I$.}{\footnotesize \par}
\end{itemize}

\textbf{\footnotesize Steps}{\footnotesize \par}
\begin{enumerate}
\item {\footnotesize Collect object hypotheses $\{o_i\}$ with local confidences $\psi^l_i$ from $I$ using a viewpoint-aware detector.}{\footnotesize \par}
\end{enumerate}

{\noindent For each object hypothesis $o_i$:}{\footnotesize \par}
\begin{enumerate}
\item[2.] {\footnotesize Define pairwise relations with the other hypotheses $o_j$ in its context $N_i$ ($o_j$ are neighboring objects of $o_i$).}{\footnotesize \par}
\item[3.] {\footnotesize Compute its contextual response $\psi^c_i = wvRN(o_i|N_i)$ via wvRN~\cite{MacskassyP07} using objects $o_j \in N_i$. \textit{(see section~\ref{sec:contextBasedPose})} }{\footnotesize \par}
\item[4.] {\footnotesize Combine the local confidence~$(\psi^l_i)$ and contextual response~$(\psi^c_i)$ to predict the object viewpoint.}{\footnotesize \par}
\end{enumerate}

\end{algorithm}

\subsection{Object Relations as Source of Context}
\label{sec:representation}
Before we discuss how relations between objects can be
used as a source of contextual information, we introduce the
representations for objects and relations used in this paper.
Given an image, we use a viewpoint-aware object detector to 
collect a set of object hypotheses $O = \{o_1 , o_2, ..., o_m \}$ 
of the categories of interest. 
Each object hypothesis $o_i$ is represented as a tuple
$o_i = (c_i, l_i, f_i, s_i)$ where $c_i$ represents the category
of the object, $l_i$ represents the location of the center of 
the bounding box of the object in the scene, $f_i$ represents 
additional object-related features (e.g. aspect ratio or scale), 
and $s_i$ the local detection score reported by the detector. 
In addition, each hypothesis is accompanied with a predicted 
discrete viewpoint $\alpha_i$. 
We will use the superscript variable $\upsilon$ on $o^{\upsilon}$ 
to indicate the state of the predicted object hypothesis. 
We refer with $o^+$ to the object hypotheses that 
are correctly localized, i.e. their predicted bounding 
boxes cover valid object instances. 
We will refer with $o^-$ to false object hypotheses.
Similarly, We use the superscript variable $\omega$ 
on $\alpha^{\omega}$ to indicate the state of the predicted viewpoint. 
We use $\alpha^+$ and $\alpha^-$ to indicate 
whether the viewpoint $\alpha$ of the object is predicted 
correctly or not. 
Finally, we will use the shorthand 
$\overline{\alpha}^\omega$ to combine the predicted viewpoint 
class and its state, i.e. $\overline{\alpha}^\omega = (\alpha,\alpha^\omega)$.

To measure the level to which an object fits in
a group of objects, we need to define a set of relations $R = \{ r_{ij} \}$
between objects. Here, we limit ourselves to pairwise
relations. 
We define these relations as the relative values between the 
attributes of the 2D bounding boxes of objects in the scene.  
Given the set of object hypotheses $O$, for each object $o_i$ 
we define pairwise relations $r_{ij}$ with each object $o_j$ 
in its neighborhood $N_i$. For simplicity, we set $N_i$ equal to the set 
composed by every other object in the image. This produces a total of \( (m (m-1))\) 
pairwise relations per image (See Figure \ref{fig:aggressiveInference}
and \ref{fig:algorithmPipeline}(c)) with $m$ the total number of 
objects in the image. 
In Section \ref{sec:implementationDetails} we describe how we 
compute the attributes that define the relations $r_{ij}$.

We summarize the notations in the following table:
\begin{table}
\centering
	\resizebox{0.4\textwidth}{!}{%
    \begin{tabular}{ | l | p{5cm} |}
    \hline
    Variable & Description \\ \hline
    $o_i = (c_i, s_i,f_i,l_i)$ &  Object hypothesis. \\ \hline
    \multirow{2}{*}{$o^\upsilon_i$}             &  Denotes whether the object hypothesis $o_i$ is true or false, $\upsilon \in (+,-)$ \\\hline
    $\alpha_i$				   &  Denotes the predicted viewpoint \\\hline
    \multirow{2}{*}{$\alpha^\omega_i$}		   &  Denotes whether $\alpha_i$ is correct or not, $\omega \in (+,-)$ \\\hline
    $\overline{\alpha}^\omega_i = (\alpha_i,\alpha^\omega_i)$ &  Shorthand \\\hline
    \multirow{2}{*}{$r_{ij}$}				   &  Relational feature computed from $(f_i,l_i)$	and $(f_j,l_j)$.\\\hline
    
    \end{tabular}}
    
    \caption{Notation summary}
\end{table}

\subsubsection{Measuring Contextual Support between Objects}
\label{sec:contextualSupport}

The problem we're addressing can be seen as a \textit{Collective Classification}
problem in which the class (viewpoint) of an object influences that of another.
We follow a simple three-step collective classification approach 
(see Alg.\ref{alg:collectiveClassification}) as proposed in \cite{MacskassyP07}.
In order to take into account the relations between objects, we estimate
a response for each object $o_i$ based on the relations 
with all the objects $o_j$ in its context. This contextual response
is obtained using the weighted-vote Relational Neighbor classifier~(wvRN) 
\cite{MacskassyP07}. This relational classifier, formally known as 
the probabilistic Relational Neighbor classifier (pRN) 
\cite{Macskassy03asimple}, is a simple, yet powerful 
classifier that is able to take advantage of the underlying structure
between networked data. It operates in a node-centric
fashion, that is, it processes one object $o_i$ at a time based
on the objects $o_j$ in its context.
During the last decade, wvRN has been successfully applied in work 
related to text mining \cite{Macskassy03asimple,MacskassyP07}, 
web-analysis \cite{MacskassyP07}, suspicion scoring 
\cite{Macskassy_suspicionscoring}, link prediction \cite{Macskassy10asunam},
and social network analysis \cite{Macskassy12icwsm,Macskassy11ICSWSM}.
More recently we applied it in the computer vision field to address
the task of context-based object pose estimation \cite{joramas:ICCV13} 
and context-based object detection \cite{joramas:WACV14}.
The \textit{wvRN} classifier \cite{MacskassyP07} computes a contextual 
score in general, as follows:

\begin{equation} 
  wvRN(o_i|N_i) = \frac{1}{Z}\sum_{o_j \in N_i} v(o_i,o_j) . w_j \\ 
  \label{eq:wvRN_ori}
\end{equation}

newith $Z=\sum w_j$ a normalization term, $v(o_i,o_j)$ a pairwise term 
measuring the likelihood of object $o_i$ given its relation
with object $o_j$, and the weighting factor $w_j$ modulating the 
effect of the neighbor $o_j$. 
In this paper we are interested in the prediction of the 
viewpoint $\alpha_i$ of each object hypothesis $o_i$.
We stress this by explicitly adding the viewpoint $\alpha_i$ in 
the equations. In addition, we apply the notation 
introduced earlier. As a result, the argument of Eq.~\ref{eq:wvRN_ori} 
is redefined as: 

\begin{equation} 
  wvRN(o_i|N_i) = wvRN(\alpha_i^+,o_i^+|N_i) 
  \label{eq:wvRN_redef}  
\end{equation}

\begin{figure*}[t!]
\centering
\includegraphics[width=0.9\textwidth]{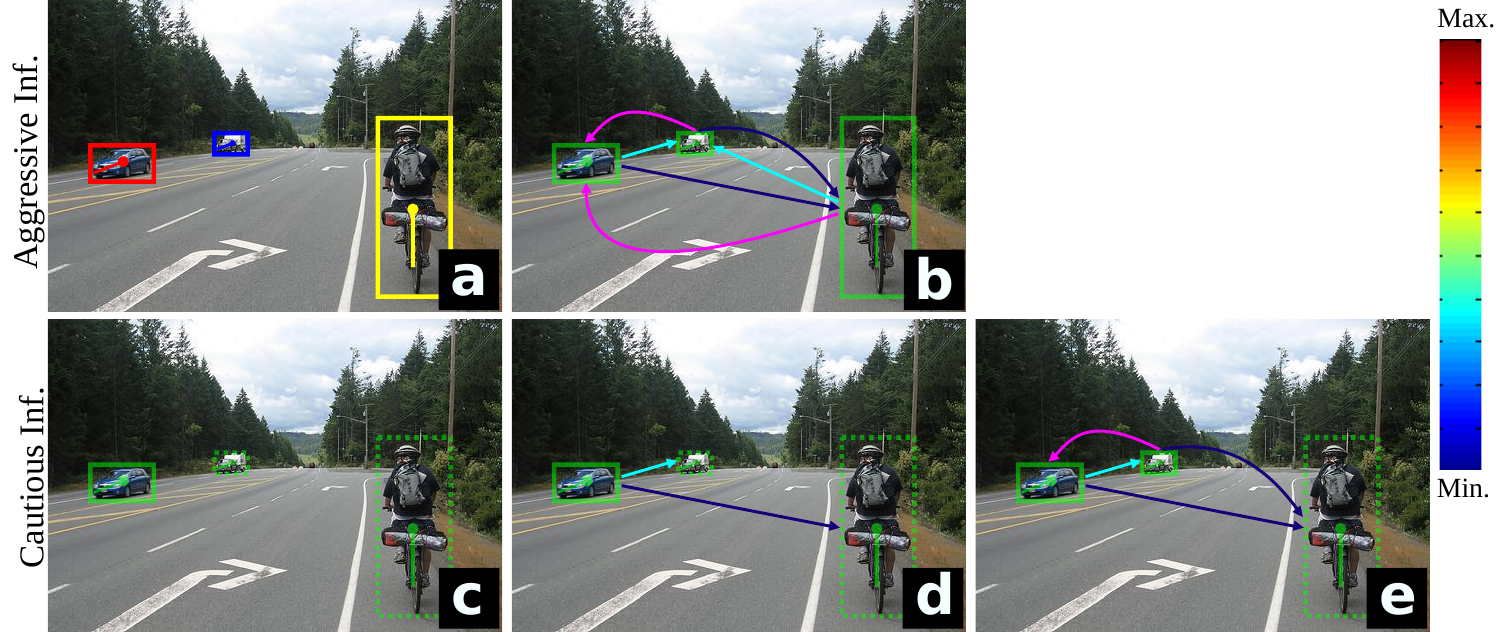}

\caption{ Types of relational inference. a) Viewpoint-aware object detection hypotheses (score encoded in jet scale), 
b) Aggressive inference, and c-e) Cautious inference. Influence of an object over another is indicated by the arrows. Solid for known objects, dashed boxes for object hypotheses to be classified. Notice how the final graph topology between b) and e) differs depending on the type of inference used.} 
\label{fig:InfType}
\end{figure*}

\subsection{Context-based Viewpoint Classification} 
\label{sec:contextBasedPose}

\noindent Given an image with a set of 2D objects $o=\{o_1,...,o_m\}$,
we estimate the viewpoint $\alpha_i$ of an object $o_i$ as the 
viewpoint $\hat{\alpha_i}$ that maximizes the likelihood of object 
$o_i$ given its neighborhood $N_i$:

\begin{equation} 
    \label{eq:contextPoseEstimation}
\hat{\alpha_i} =  \arg\max_{\alpha_i} (~ wvRN(\overline{\alpha_i}^+,o_i^+|N_i) ~),	\\    
\end{equation}

As mentioned earlier, the group fitting of an object 
is measured by the output of the \textit{wvRN} classifier \cite{MacskassyP07} 
which is defined, for our specific task, as follows:

\begin{equation}
  \label{eq:wvRN}
wvRN(\overline{\alpha_i}^+,o_i^+|N_i) = \frac{1}{Z}\sum_{o_j \in N_i} p(\overline{\alpha_i}^+, o_i^+|r_{ij},c_i) \cdot w_j	\\
\end{equation}

In our formulation $w_j$ is a weighting term that takes 
into account the noise in the object detector (see below). 
The original pairwise term $v(o_i,o_j)$ is defined as
\(p(\overline{\alpha_i}^+,o_i^+|r_{ij},c_i)\). This conditional represents 
the probability of object \(o_i\), of category $c_i$, being a true 
hypothesis $o_i^+$, with correctly predicted viewpoint $\overline{\alpha_i}^+$, 
given its relation $r_{ij}$ with object \(o_j\). 
Using Bayes' Rule we estimate \(p(\overline{\alpha_i}^+,o_i^+|r_{ij},c_i)\) 
as the posterior:

\begin{multline}
            p(\overline{\alpha}_i^+,o_i^+|r_{ij},c_i) = \frac {p(r_{ij}|\overline{\alpha}_i^+,o^+_i,c_i)p(\overline{\alpha}_i^+,o^+_i|c_i)}{ p( r_{ij} | c_i ) } \\ \\ = \frac {p(r_{ij}|\overline{\alpha}_i^+,o^+_i,c_i)p(\overline{\alpha}_i^+,o^+_i|c_i)} { \sum\limits_{\upsilon~\in~ \{+,-\}} \sum\limits_{ \omega~\in~\{+,-\}} ~ p(r_{ij},\overline{\alpha}_i^\omega,o^\upsilon_i|c_i) }\\ \\ = \frac {p(r_{ij}|\overline{\alpha}_i^+,o^+_i,c_i)p(\overline{\alpha}_i^+,o^+_i|c_i)}{\sum\limits_{\upsilon~\in~ \{+,-\}} \sum\limits_{ \omega~\in~\{+,-\}} ~ p(r_{ij}|\overline{\alpha}^\omega_i,o^\upsilon_i,c_i)p(\overline{\alpha}^\omega_i,o^\upsilon_i|c_i) }
   \label{eq:relationProb}   
\end{multline}



The components of Eq.\ref{eq:relationProb} are obtained through 
the following procedure. During the training stage, we compute 
pairwise relations $r_{ij}$ between the annotated objects in the 
training images. Furthermore, we extend this set of objects and 
relations by running a local detector on the training set 
producing a set of hypotheses per image. Then, we flag the 
hypotheses as true positive hypotheses $o_i^+$ or as 
false positive hypotheses $o_i^-$ considering spatial matching 
based on the Pascal VOC \cite{PascalVOC2012} matching criterion. 
In addition, each object hypothesis is flagged as $\overline{\alpha}_i^+$ 
or $\overline{\alpha}_i^-$ depending on whether its viewpoint was predicted 
correctly or not, respectively.
Note that hypotheses with label combination $(\overline{\alpha}_i^+,o_i^-)$ 
do not exist since it is not possible to predict correctly the 
viewpoint of a false hypothesis.
In order to avoid repeated object instances, we replace true 
hypotheses $o_i^+$, with correctly predicted viewpoint $\overline{\alpha}_i^+$, 
by their corresponding annotations. Similarly, we replace the 
relations produced by these correct hypotheses by those produced 
by their corresponding annotations. 
This step of integrating the hypotheses in the training data, 
allows our method to model, up to some level, the noise in 
the relations $r_{ij}$ introduced by the local detector. \newtext{More 
specifically, this noise appears in the form of frequent false relations 
that arise from common false hypotheses that may be predicted 
by the local detector.}
This produces a set of objects $o_i$ with their corresponding 
pairwise relations $R=\{r_{ij}\}$ from the whole training set. 
\newtextRed{
Using this information, we estimate a probability density 
function (pdf) via Kernel Density Estimation (KDE).
Finally, during testing, \(p(r_{ij}|\overline{\alpha}_i^+,o^+_i,c_i)\), 
\(p(r_{ij}|\overline{\alpha}_i^-,o_i^+,c_i)\) and 
\(p(r_{ij}|\overline{\alpha}_i^-,o_i^-,c_i)\) are computed by 
evaluating the pdf at the test points defined by the 
relations $r_{ij}$ computed between object hypotheses.} 

This method captures the statistics of typical configurations. 
The priors $p(\overline{\alpha}_i^+,o_i^+|c_i)$, $p(\overline{\alpha}_i^-,o_i^+|c_i)$ and 
$p(\overline{\alpha}_i^-,o_i^-|c_i)$ are estimated per object category based 
on their occurrence in the training set, respectively.
Moreover, given that hypotheses with label combination $(\overline{\alpha}_i^+,o_i^-)$ 
do not exist, we set the prior $p(\overline{\alpha}_i^+,o_i^-|c_i) = 0$.
Note that the conditional terms $p(r_{ij}|\overline{\alpha}^\omega_i,o^\upsilon_i,c_i)$  
are derived from a distribution representing the relational space 
covering pairwise relations from specific object categories. 
For this reason, in order to consider relations between objects 
of different categories, during training, we estimate the pdfs
from pairwise relations (samples) with a specific object category, 
as source, and a specific object category, as target. Given a set of 
$n$ object categories of interest, we model a total of $n^2$ 
pdfs that will be used later to compute the term 
\(p(r_{ij}|\overline{\alpha}^\omega_i,o^\upsilon_i,c_i)\).

The weighting factor $w_j$ of Eq.~\ref{eq:wvRN} takes into account 
the noise that is introduced by the object detector in the predicted 
neighboring objects $o_j$. We estimate $w_j$ using a \textit{Probabilistic 
Local Classifier} that takes into account the score $s_j$ provided by the 
object detector for its respective hypothesis $o_j$. The output of this 
classifier will be the posterior $p(\overline{\alpha}_j^+,o_j^+|s_j,c_j)$ of 
object $o_j$ of category $c_j$ being correctly localized ($o_j^+$), with 
correctly predicted viewpoint $\overline{\alpha}_j^+$, given its score $s_j$. 
We compute this posterior following the procedure presented in 
\cite{perkoCVIU2010}:

\begin{equation}
\label{eq:probLocalClassifier}
\begin{aligned}
  w_j &= p(\overline{\alpha_j}^+,o^+_j|s_j,c_j) \\ \\
      &= \frac {p(s_j|\overline{\alpha_j}^+,o^+_j,c_j)p(\overline{\alpha_j}^+,o^+_j|c_j)}
  {\sum\limits_{\upsilon~\in~\{+,-\}} \sum\limits_{\omega~\in~\{+,-\}} ~ p(s_j|\overline{\alpha}^\omega_j,o^\upsilon_j,c_j)p(\overline{\alpha}^\omega_j,o^\upsilon_j|c_j) }
\end{aligned}  
\end{equation}

The components of this equation are obtained following 
a procedure similar to that for Eq.~\ref{eq:relationProb} 
up to the point where hypotheses are assigned the labels 
$\overline{\alpha}^+$, $\overline{\alpha}^-$, 
$o^+$ and $o^-$. Then, based on these 
flagged hypotheses, we compute the conditionals 
$p(s|\overline{\alpha}^+,o^+,c)$, $p(s|\overline{\alpha}^-,o^+,c)$ and 
$p(s|\overline{\alpha}^-,o^-,c)$ respectively via KDE. 
Finally, the priors $p(\overline{\alpha}^+,o^+|c)$, 
$p(\overline{\alpha}^-,o^+|c)$ and $p(\overline{\alpha}^-,o^-|c)$ 
are estimated per category as the corresponding proportions 
of labeled hypotheses in the training data. 
As a result, $p(\overline{\alpha}_j^+,o_j^+|s_j,c_j)$ 
expresses the probability of a hypothesis being correct 
given its detection score. This procedure allows us to 
plug-in any standard object detector in our method. 
Note that, similar to the term $p(r_{ij}|\overline{\alpha}_i^+,o^+_i,c_i)$, 
the conditional $p(s|\overline{\alpha}^+,o^+,c)$ 
is computed from object category-wise KDE, 
where all the sample points are derived from 
object hypotheses belonging to the object category 
of interest.

\subsection{Cautious Inference}
\label{sec:cautiousInf}
Following the definition introduced in \cite{McDowellCautiousInfGA07}, an 
algorithm is considered ``cautious'' if it seeks to identify and employ 
the more certain or reliable relational information. 
We focus on two factors that \cite{McDowellCautiousInfGA07} introduced to 
control the degree of caution in an algorithm. The first factor dictates 
to use only objects for which the prediction is confident enough. 
The second factor increases caution by favoring already-known 
relations. These are relations that have been seen in 
the training images.

For the aggressive version of our relational classifier, we use wvRN 
as described in Eq.~\ref{eq:wvRN}. For each object hypothesis to be
classified, it considers \textit{all} the other objects $o_j$ in its 
context $N_i$ during the inference (see Figure \ref{fig:InfType}(b)). 
For the \textit{cautious} version of our relational classifier, we 
enforce the above principles in the following fashion. 

For the first principle, giving relevance 
to the most-certain objects, we perform an iterative approach inspired
by \cite{nevilleJensenIterativeInf}. Given a set of hypotheses 
\( O = \{o_1,...,o_n\} \), we define the disjoint sets $O^k$ and 
$O^u$ containing the known and unknown objects, respectively, with
$ O = O^k \cup O^u$ at all times. During inference, we initialize 
$O^k = \{\}$ and $O^u=O$ and flag as \textit{known} object, the 
hypothesis with the highest score based on the probabilistic local
classifier (Eq.~\ref{eq:probLocalClassifier}) . 
This hypothesis is moved to the set of known objects $O^k$. Then,
the wvRN score for each of the unknown objects $o_i \in O^u$ is re-estimated 
considering \textit{only} the known objects $o_j \in O^k$ in their 
context $N_i$. This redefines Eq.~\ref{eq:wvRN} in the 
following way:

\begin{equation}
\label{eq:ite_wvRN}
wvRN(\overline{\alpha}_i^+,o^{+}_i|N_i) = \frac{1}{Z}\sum_{\mathclap{~o_j \in (N_i \cap O^k)}} p(\overline{\alpha}_i^+,o^{+}_i|r_{ij},c_i) \cdot w_j\\ 
\end{equation}

We flag the hypothesis with highest wvRN response as \textit{known} and 
move it to the set of known objects $O^k$. We repeat this procedure 
promoting one hypothesis $o_i \in O^u$ at a time until the set of unknown 
objects $O^u$ is empty. Finally, for the sake of similarity in the 
ranking of the new scores, we re-estimate the score of the first 
promoted object using Eq.~\ref{eq:ite_wvRN} with the second promoted 
object as known contextual object.

For the second principle of \textit{cautious} inference: ``favoring 
relations already seen on training data'', our use of 
KDE for estimating the vote $p(\overline{\alpha}_i^+,o^{+}_i|r_{ij},c_i)$ 
from each contextual object $o_j$ implicitly introduces this 
characteristic in the inference. 
 
For the sake of clarity, we illustrate cautious inference with an
example. Consider the hypotheses provided by a viewpoint-aware
detector shown in Figure \ref{fig:InfType}(a). Note that their detection
score is encoded in jet scale, giving the hypothesis in red a higher 
score than the one in blue. Since there are three object hypotheses,
there will be three steps during cautious inference.
During the first step, the hypothesis in red is promoted as known object
(Figure~\ref{fig:InfType}(c)) making it a valid source of contextual 
information for the others (Figure~\ref{fig:InfType}(d)). 
During the second step, the hypothesis initially in blue, with higher 
relational score, is promoted as known object. Again, this makes this 
hypothesis a source of context for the remaining hypotheses. In 
addition, this second promoted hypothesis will be used to re-estimate 
the first one. Finally, the last, initially yellow, hypothesis is 
estimated by using all the known hypotheses as context 
(Figure~\ref{fig:InfType}(e)).

\subsection{Combining Local and Contextual information} 
\label{sec:responseCombination}

At this point, we have gathered a set of object hypotheses 
$O = \{o_1 , o_2, ..., o_m \}$ using a viewpoint-aware object detector. 
For each hypothesis $o_i$ we have, on the one hand, its local response
$\psi^l_i$ which consists of the viewpoint $\alpha_i$ and score $s_i$ reported by 
the object detector based purely on local features. 
On the other hand, we have its contextual response $\psi^c_i$ defined
by the relational response $wvRN(\overline{\alpha}_i^+,o_i^+|N_i)$ (Eq.~\ref{eq:wvRN}) over 
different viewpoints.
These two responses $\psi^l$ and $\psi^c$ have complementary behaviors. While 
the local response $\psi^l$ pulls the decision towards intrinsic object 
features, the contextual response $\psi^c$ pulls the decision in such
a way that the object to be classified fits in the group of objects
in the image. In order to find a balance between
these responses, for each hypothesis $o_i$ we build a coupled-response 
vector $\Psi_i=[ \psi^l_i , \psi^c_i ]$ and estimate the viewpoint $\hat{\alpha_i}$ 
of the object as:

\begin{equation} 
    \label{eq:combinedPoseEstimation}  
\hat{\alpha_i} =  \arg\max_{\alpha_i} (~f(\overline{\alpha}_i^+|\Psi_i)~),	\\    
\end{equation}

where $f$ is a multiclass classifier trained from coupled-response 
vector - viewpoint annotation pairs $(\Psi,\alpha)$ extracted from object 
hypotheses collected from a validation set. In Section~\ref{sec:implementationDetails} 
we give more details about the multiclass classifiers used in 
our experiments.

\section{Implementation Details}
\label{sec:implementationDetails}

\subsection{Object Detection}
Since the main focus of this work is on the task of object viewpoint estimation
we will leave the specific task of localizing/detecting the objects of 
interest, based on intrinsic features, to an off-the-shelf detector. 
We selected detectors that not only provide the localization 
(bounding box) of the object but also a viewpoint prediction discretized 
into 8 viewpoints. 

In this work we use three different viewpoint-aware detectors,  
two of which are variations of the deformable part-based model detector (DPM) 
\cite{felzenszwalbTPAMI2009}, where a specific component of the model 
is learned for each of the discrete viewpoints to be classified. 
In particular, we use the \textit{mDPM} detector proposed by Lopez et al. \cite{rlopez_2011}, 
and the \textit{LSVM-MDPM-sv} detector from Geiger et al. \cite{GeigerNIPS11}.
\newtext{The third detector, \textit{Faster RCNN - viewpoint CNN}, is based on state of the art learning-based representation methods implemented via convolutional neural networks (CNN). It is composed of a faster RCNN detector \cite{ren15fasterrcnn}, used to localize object instances, combined with a fine-tuned CNN Alexnet architecture \cite{alexnetNIPS12} to classify the viewpoint of the predicted object bounding boxes.}

\subsection{Pairwise Relations Extraction}
\label{sec:pairwiseRelations}
\begin{figure}[t!]
\centering
\includegraphics[width=0.5\textwidth]{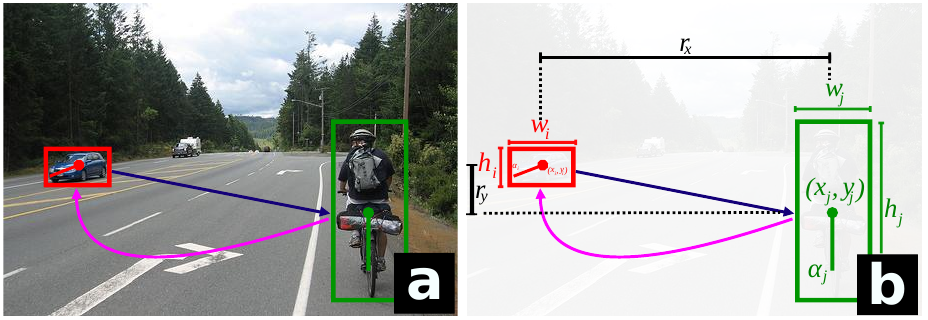}

\caption{ Attributed pairwise relations defined from object hypotheses predictions. $w_i$: bounding box width, $h_i$: bounding box height, $(x_i,y_i)$: coordinates of
the center of the bounding box, and $\alpha_i$: object viewpoint.} 
\label{fig:relationAttributes}
\end{figure}

Given a set of objects in the scene, we define 
pairwise relations by deriving relative attributes from
the bounding boxes that cover the objects. Different
from \cite{joramas:ICCV13}, the objects are 2D entities projected in the image space. 
Given a set of objects $O=\{o_1,o_2,...,o_m\}$, for each of the 
objects \(o_i\) , we measure the relative location $(rx_{ij},ry_{ij})$,
relative scale $rs_{ij}$ and viewpoint $\alpha_{j}$ of 
each of the other objects \(o_j\), producing a relational descriptor 
\( r_{ij} = (rx_{ij},ry_{ij},rs_{ij},\alpha_{j}) \), see Figure~\ref{fig:relationAttributes}. 
We define the relative attributes of the pairwise relations 
in the following as:
$rx_{ij}=(\frac{x_j-x_i}{w_i})$, $ry_{ij}=(\frac{y_j-y_i}{h_i})$ and
$rs_{ij}=(\frac{w_j}{w_i},\frac{h_j}{h_i})$, where $(x_i,y_i,w_i,h_i)$  
define the center, width and height of the bounding box of object $o_i$.
This produces pairwise relations defined by five attributes.
\newtextRed{The number of pairwise relations per image has a quadratic growth w.r.t.
the number of objects}, more precisely, for an image with \textit{m}
objects a total of \( (m (m-1))\) pairwise relations are extracted.

\subsection{Multiclass Classification for coupled object \\viewpoint estimation}
\label{sec:responseCombinationDetails}
In order to enforce consistency between the local response $\psi^l$, given
by the detector, and the contextual response $\psi^c$, given by the context-based
viewpoint classifier, we define object viewpoint classification as a classification
problem (Eq.~\ref{eq:combinedPoseEstimation}) based on the coupled-response $\Psi=(\psi^l,\psi^c)$.
In this paper we evaluate the performance of  two methods for this particular
task. 

The first method, \textit{Probabilistic Combination}, is inspired by \cite{perkoCVIU2010}. As 
Eq.~\ref{eq:OKDEcomb} presents, to classify the viewpoint 
$\hat{\alpha_i}$ of an object $o_i$ we perform {\small MAP} inference for 
the coupled-response $\Psi_i$ over the discrete viewpoint classes $\alpha_k$.

\begin{equation} 
\label{eq:OKDEcomb}
\begin{aligned}
 \hat{\alpha_i} &= \arg\max_{\alpha_k} (~p(\overline{\alpha}_k^+|\Psi_i)~) \\
                &= \arg\max_{\alpha_k} (~p(\Psi_i|\overline{\alpha}_k^+)p(\overline{\alpha}_k^+)~)
\end{aligned}
\end{equation}

In this equation, the term $p(\Psi_i|\overline{\alpha}_k^+)$ is estimated using KDE from 
the coupled-responses $\Psi=[\psi^l,\psi^c]$ of object hypotheses collected from a 
validation set. The term $p(\overline{\alpha}_k^+)$ is determined by the proportion of
objects with viewpoint $\alpha_k$ in the validation set.

The second method, \textit{Linear Combination}, is based on linear Support Vector Machines (SVM).
Specifically we use the method from Crammer and Singer~\cite{crammerJMLR2001} 
for the implementation of multiclass SVMs. Following this procedure, 
the object viewpoint classification problem is defined as:

\begin{equation} 
    \label{eq:SVMcomb}
    \hat{\alpha_i} = \arg\max_{\alpha_k}  ( W_k \cdot \Psi_i).
\end{equation}

Here the problem is to learn the matrix of weights $W_k$ of
the SVM Model that will be used to predict the class, in this 
case the object viewpoints $\alpha_k$.
Similar to the previous classifier, here we collect the response vectors
$\Psi_i$ from validation images. In addition, we perform 3-fold cross validation
to estimate the cost parameter used for training the SVM classifier.

\subsection{Kernel Density Estimation}

Kernel Density Estimation (KDE) is performed using the odKDE variant 
proposed in \cite{KristanODKDE}. Under odKDE the sample distribution 
is modeled by a $C$-component mixture of Gaussians as
\(f(x) = \sum_{i=1}^C{ \omega_i \cdot \phi_{\Sigma_{si}}( \ x-\mu_i )  } \), 
where $\phi_{\Sigma_{si}}$ is a Gaussian centered in $\mu_i$ and with covariance matrix 
$\Sigma_{si}$. The effect of each component is measured by 
the mixture weights $\omega_i$ with $\sum{\omega_i}=1$. As Eq.\ref{eq:ODKDE} shows, the KDE $p(x)$ 
of the distribution is determined by the convolution of the sample 
distribution $f(x)$ with a Gaussian kernel $\phi_H (x)$.

\begin{equation} 
\label{eq:ODKDE}
\begin{aligned}     
    p(x) &= \phi_H (x) * f(x) \\
         &= \sum_{i=1}^C{ \omega_i \cdot \phi_{\Sigma_i}( \ x-\mu_i )  } 
\end{aligned}
\end{equation} 

where $\Sigma_i = H_i + \Sigma_{si}$, and $H_i$ is
the bandwidth of the convolution kernel. For more details related
to the computation of $H_i$ please refer to \cite{KristanODKDE}.
In our experiments, the samples $x$ take the form of the attributes
of the pairwise relations $r_{ij}$ between objects 
(Eq.~\ref{eq:relationProb}), the detection scores $s_i$ 
(Eq.~\ref{eq:probLocalClassifier}) or the coupled-response $\Psi_i$
(Eq.~\ref{eq:OKDEcomb}). In these equations, KDE is used to 
estimate the value of their respective conditional terms.

\begin{table*}[ht!]

\centering

\begin{tabular}{|l|c|c||c|c|}
\hline

 \multicolumn{5}{|l|}{ \textbf{KITTI dataset \cite{Geiger12KITTI}}}  \\ \hline 
 & \multicolumn{2}{|c||}{ \textbf{MPPE}} & \multicolumn{2}{|c|}{ \textbf{AVP}}  \\ \hline 
 \textbf{Method} & \textbf{Oracle}  &  \textbf{Lopez et al. \cite{rlopez_2011}} & \textbf{Oracle}  &  \textbf{Lopez et al. \cite{rlopez_2011}}  \\  
  Aggressive-RF1 & 0.37 & 0.27 & 0.74 &	0.06 \\
  Aggressive-RF2 & 0.22 & 0.17 & 0.60 &	0.06 \\
  Cautious-RF1   & 0.41 & 0.30 & 0.72 &	0.06 \\
  Cautious-RF2   & 0.36 & 0.27 & 0.66 &	0.07 \\
  
\hline   
\textbf{Method} & \textbf{Oracle}  &  \textbf{Geiger et al. \cite{GeigerNIPS11}} & \textbf{Oracle}  &  \textbf{Geiger et al. \cite{GeigerNIPS11}}  \\   
  Aggressive-RF1 & 0.39 & 0.28 & 0.68 &	0.10 \\
  Aggressive-RF2 & 0.23 & 0.20 & 0.55 &	0.14 \\
  Cautious-RF1   & 0.40 & 0.32 & 0.69 &	0.09 \\
  Cautious-RF2   & 0.36 & 0.31 & 0.64 &	0.10 \\
  
\hline 
 \textbf{Method} & \textbf{Oracle}  &  \textbf{Ren et al. \cite{ren15fasterrcnn}} & \textbf{Oracle}  &  \textbf{Ren et al. \cite{ren15fasterrcnn}}  \\  
  Aggressive-RF1 & 0.48 & 0.36 & 0.78 &	0.28 \\
  Aggressive-RF2 & 0.28 & 0.29 & 0.60 &	0.25 \\
  Cautious-RF1   & 0.43 & 0.44 & 0.74 &	0.30 \\
  Cautious-RF2   & 0.39 & 0.40 & 0.67 &	0.28 \\

\hline

\hline 
\end{tabular}

\caption{ Contextual object viewpoint classification 
mean precision for pose estimation (MPPE)  
and Average viewpoint precision (AVP) for the proposed 
methods on the KITTI \cite{Geiger12KITTI} dataset.
We report results on hypotheses collected with
DPM-based methods (\cite{rlopez_2011,GeigerNIPS11}) 
and a CNN-based method (Faster RCNN~\cite{alexnetNIPS12} 
+ viewpoint CNN~\cite{ren15fasterrcnn}).
Note how the methods that perform \textit{cautious} 
relational inference tend to have superior performance 
than their \textit{aggressive} counterparts.}

\label{table:contextClassificationKITTI}

\end{table*}


\section{Evaluation } 
\label{sec:evaluation}

\subsection{Experimental Details}
\label{sec:experimentalDetails}

\textbf{Datasets:}
We focus on urban scenes. For this reason, we 
conduct experiments on the object detection 
set of the KITTI benchmark \cite{Geiger12KITTI}.
%
The KITTI dataset is collected from a car-mounted camera, 
resembling an autonomous navigation setting. We consider ``car''
as category of interest as it occurs multiple times in  each
image of this dataset. 
This dataset presents a variety of difficult scenarios ranging from object 
instances with high occlusions to object instances with very small 
size. Furthermore, it provides precise annotations from objects in 
the 2D image and in the 3D space, including their respective viewpoints.
Since this is a benchmark dataset, annotations are not available 
for the test set. For this reason, we focus our experiments on the 
training set. Each image of the training set belongs to a 
video sequence; we use the sequence ID and the time stamp of 
each image in order to sort them by video sequence. Then, 
the training images belonging to each sequence are split in 
three disjoint subsets in chronological order.
The first subset is used for 
learning the relations between object instances 
(see Section~\ref{sec:contextBasedPose}). The second subset 
is used to learn the combination of local and contextual 
information (see Section~\ref{sec:responseCombination}). The third 
set is used to evaluate the performance of the proposed method.

Since the focus of this work is on reasoning about relations 
between objects, we focus our evaluation on the subset of 5266 
images with two or more object instances.

\vspace{2mm}
\noindent\textbf{Methods:}
In our experiments, we define four methods to perform 
context-based classification 
based on the combination of the following parameters. 
As presented in Section \ref{sec:contextBasedPose} our 
methods are defined as \textit{Aggressive} or \textit{Cautious} 
depending on the type of relational inference they 
perform. In addition, we define pairwise relations following 
two formats. The first format, $RF1$, is as defined in
Section \ref{sec:pairwiseRelations}. The second format, 
$RF2$, is similar to $RF1$ but removing the viewpoint 
attribute $\alpha_j$, thus producing a relation 
\( r_{ij} = (rx_{ij},ry_{ij},rs_{ij}) \) with 
four attributes that encodes only relative location 
and scale.

\subsection{Contextual Object Viewpoint Classification }
The objective of this first experiment is to evaluate the performance 
of the algorithm at estimating the viewpoint of the object hypotheses
purely based on its neighboring objects 
(see Section \ref{sec:contextBasedPose}). 

We evaluate our methods following two settings to collect object 
hypotheses. The first setting starts from an ``Oracle'' detector,  
thus producing perfectly localized hypotheses. Furthermore, in this 
oracle setting, while the viewpoint of an object is being classified, the 
ground truth viewpoints of its neighbor objects are used. 
This setting will show whether there is something to gain from 
reasoning about object relations for object viewpoint classification.

The second setting uses the previously mentioned viewpoint-aware detectors 
(\cite{GeigerNIPS11,rlopez_2011,ren15fasterrcnn,xiang_wacv14}) to collect object 
hypotheses. 
Then, the viewpoint of each object is estimated based on its 
neighbors taking into account the detection scores and predicted viewpoints. 
This last setting represents a more realistic scenario.

In this experiment we will use the Mean Precision for 
Pose Estimation (MPPE) and Average Viewpoint Precision (AVP) 
as performance metrics. The MPPE metric is 
traditionally used to measure viewpoint classification 
\cite{LS10,rlopez_2011,PepikCVPR12,savarese:iccv2007}. 
MPPE is computed as the average of the diagonal of the 
class-normalized confusion matrix of the viewpoint classifier.
\newtext{
Additionally, taking into account the observations made by \cite{RendondoECCV16}, 
we include AVP \cite{xiang_wacv14} as an additional performance metric. AVP a is metric derived from Average Precision (AP), traditionally used to measure object localization performance, extended to measure the capability of a viewpoint-aware 
detector at predicting the location and viewpoint of object instances, jointly.
We report the results of this experiment in Table~\ref{table:contextClassificationKITTI}.
}

\textbf{Discussion:}
Notice first that when inspecting the MPPE performance obtained using 
an \textit{Oracle} detector all the performance values are above 
chance levels ($\sim$0.13 for classification of 8 discrete viewpoints). 
This suggests that indeed there is some information about the 
viewpoint of the unknown object that can be gathered from its 
neighboring objects - hence it makes sense to reason about object 
relations for object viewpoint estimation.
We see a similar trend for the methods that collect object hypotheses 
using one of the viewpoint-aware detectors 
(\cite{GeigerNIPS11,rlopez_2011,ren15fasterrcnn,xiang_wacv14}), 
but with a drop in performance. This is to be expected, since 
some of the object hypotheses used as sources of context are false 
positives which introduce noise in the inference. 
We can also see that the performance increases as we go to the 
CNN-based method. Given the fact that between the \textit{Oracle}-based 
methods the only difference is the learned relational model used to compute 
the pairwise term $v(o_i,o_j)$ (Section \ref{sec:contextualSupport} 
and \ref{sec:contextBasedPose}) and that this model depends on the 
annotations and hypotheses being collected by the detector, 
it is evident that using a better performing detector can help 
improving the modeling of context and hence improves viewpoint 
classification.

\newtext{
We can notice that the values of MPPE are significantly lower 
when compared to their respective AVP counterparts. This is to be 
expected since AVP is a harder metric which also measures the level 
to which an object hypothesis is properly localized. Furthermore, 
we notice that AVP values for the CNN-based methods are superior again 
to those of the DPM-based methods. This confirms the known fact 
that DPM-based methods, while being able to perform viewpoint 
classification to some extent, have a limited capability at properly 
localizing object instances as compared to the CNN-based methods. 
}

Regarding relational inference type, we notice that in all cases 
cautious relational inference has superior performance than 
aggressive inference. In the oracle 
setting, methods based on cautious inference outperform their aggressive 
counterparts by 7 percentage points (pp). In the realistic setting, 
this difference increases to 8 pp. This is probably caused by the extra 
room for improvement that is produced by noise in standard detectors.

Finally, related to the format used to represent the pairwise 
relations, it seems that relations defined by \textit{RF1} 
(i.e. including the pose of the contextual objects) have superior 
performance over \textit{RF2}. In purely contextual classification, 
relations defined by \textit{RF1} outperform those defined by 
\textit{RF2} by 11 and 12 pp, in the ideal and realistic setting, 
respectively. 
\newtext{
For the case of joint localization and viewpoint 
prediction, on the one hand, we notice that the performance of 
\textit{Oracle} baselines are led by methods based on \textit{RF1}-type relations 
with an average difference of 16 AVP points. 
Keeping in mind that \textit{RF1} relations are similar to \textit{RF2} relations 
with the difference that \textit{RF1} includes the viewpoints of the 
contextual instances, this suggests that considering viewpoint 
information from other object instances is an informative 
contextual cue for the problem at hand. 
On the other hand, we notice that when starting from the hypotheses 
collected with the viewpoint-aware detectors, the difference between 
\textit{RF1} and \textit{RF2} is not so outspoken. In fact, for the case of the 
DPM-based methods, which have noisier viewpoint predictions, this 
difference is almost nonexistent or reverted.
For the case of the CNN-based method, \textit{RF1}-type relations still 
outperform their \textit{RF2} counterparts with 3 AVP points on average. 
Finally, compared to the DPM-based methods, the difference in 
performance between the \textit{Oracle} and the real detectors 
is more reduced for the CNN-based methods.
}


\begin{figure*}
\centering
\includegraphics[width=1\textwidth]{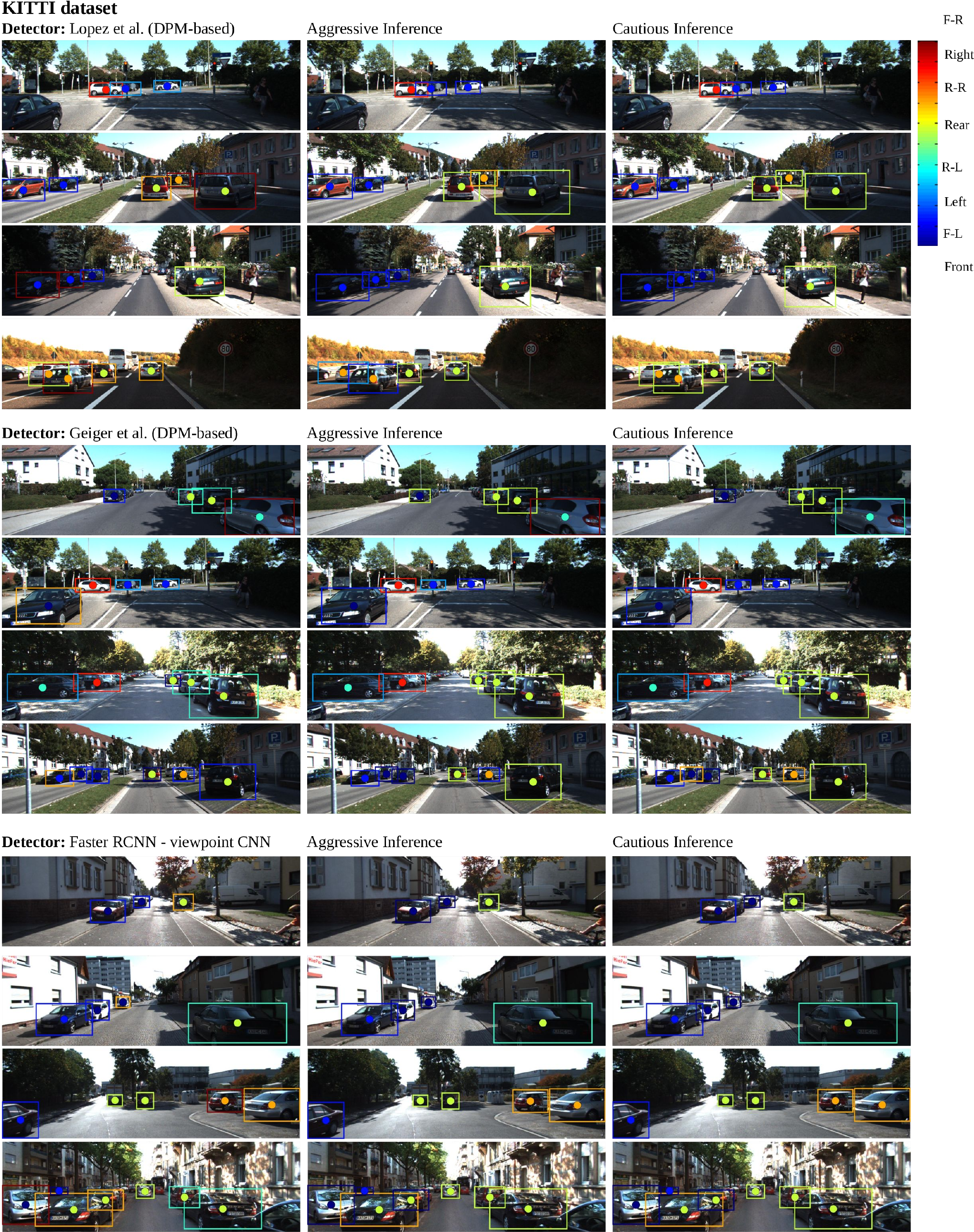}

\caption{ Context-based viewpoint classification qualitative results for 
cars in the KITTI \cite{Geiger12KITTI}. Object viewpoint classification 
results are encoded in jet scale (see color bar).
Continuous line, predicted object viewpoint;  Circle, ground-truth object 
viewpoint (Best viewed in color).}

\label{fig:qualitativeResults}
\end{figure*}


\begin{figure*}
\centering
\includegraphics[width=1\textwidth]{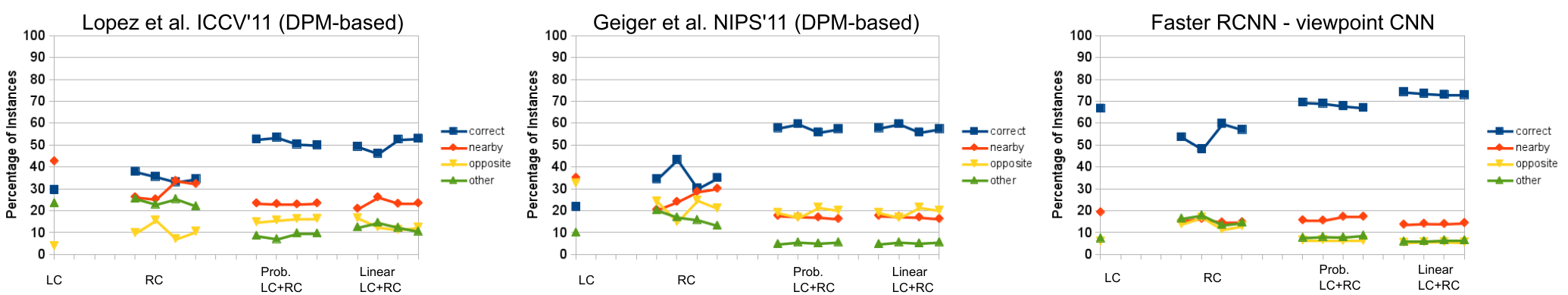}

\caption{Viewpoint classification error. For each of the baselines we 
show the percentage of object instances whose viewpoint is: predicted 
correctly (blue), confused with a nearby viewpoint class (red), 
confused with an opposite viewpoint (yellow), and confused with other 
viewpoints (green). We report results for both the local methods (LC)
which only consider local information, the relational methods (RC) which 
only consider contextual information and the probabilistic and linear 
combination of both cases. For the methods based on contextual information 
we report results for four variants \textit{Aggressive-RF1}, \textit{Aggressive-RF2}, 
\textit{Cautious-RF1} and \textit{Cautious-RF2}.
It is noticeable that the proposed context-based methods assist local 
methods by addressing \textit{nearby} and \textit{other} type errors.}

\label{fig:viewpointErrorAnalysis}
\end{figure*}

\subsection{Combining Local and Contextual cues for Viewpoint Estimation }
In this experiment we measure the performance of the
combination of local and contextual information for 
object viewpoint estimation. Specifically, we evaluate 
the late fusion of the responses from the local classifier 
(LC), i.e. the viewpoint-aware detector, and the relational classifier 
(RC), i.e. the viewpoint classifier based on the context 
(see Sections~\ref{sec:responseCombination} \& 
\ref{sec:responseCombinationDetails}). 
The objective of this experiment is to verify whether 
enriching the local classifier with contextual 
information can increase the performance initially 
obtained by the local classifier alone. 
For this experiment we use the same MPPE and AVP performance 
metrics as in the previous experiment.
We report quantitative results of this experiment in Tables 
\ref{table:combinedSingleClassExperiment} and 
\ref{table:combinedSingleClassExperimentAVP}. See Figure 
\ref{fig:qualitativeResults} for some qualitative results.

\begin{table}[h!]
\centering
\begin{tabular}{|l|c|c|c|c|c|c}
\cline{1-6}
\multicolumn{6}{|l|}{ \textbf{KITTI dataset \cite{Geiger12KITTI} } - MPPE performance } \\
\cline{1-6} 
\multirow{2}{*}{\backslashbox{\scriptsize{LC}}{\scriptsize{RC}} }
		     &  \multirow{2}{*}{None}    & \multicolumn{2}{c|}{Aggressive} & \multicolumn{2}{|c|}{Cautious}  \\ 
		     & & \footnotesize{RF1} & \footnotesize{RF2} & \footnotesize{RF1} & \footnotesize{RF2} \\
\cline{1-6} \cline{1-6}
 None		     	       	 & -    & 0.27 & 0.17 & 0.30 & 0.27 \\
\multirow{2}{*}{Lopez et al.\cite{rlopez_2011} }  & \multirow{2}{*}{0.32} & 0.37 & \textbf{0.41} & 0.34 & 0.37 & \scriptsize{Prob.}\\
						  &  & 0.28 & 0.35 & 0.41 & \textbf{0.44} & \scriptsize{Linear} \\

\hline\hline
None		     		 & -    & 0.28 & 0.20 & 0.32 & 0.31 \\
\multirow{2}{*}{Geiger et al.\cite{GeigerNIPS11}} & \multirow{2}{*}{0.44} & 0.39 & \textbf{0.45} & 0.40 & 0.41 & \scriptsize{Prob.} \\   
				 &   & 0.39 & \textbf{0.45} & 0.36 & 0.39 & \scriptsize{Linear} \\   
				 
\hline\hline
				 
None		     		 & -    & 0.36 & 0.29 & 0.44 & 0.40 \\
\multirow{2}{*}{Ren et al.\cite{ren15fasterrcnn}} & \multirow{2}{*}{\textbf{0.61}} & 0.56 & 0.53 & 0.51 & 0.50 & \scriptsize{Prob.} \\   
				 &   & 0.59 & 0.59 & 0.59 & 0.58 & \scriptsize{Linear} \\   				 				 
				 
\cline{1-6}     

\end{tabular} 

\caption{ Combined object viewpoint classification performance on 
KITTI \cite{Geiger12KITTI}. 
Mean Precision on Pose Estimation (MPPE) performance is presented for the combination of 
the local classifier (LC) and relational classifier (RC) 
using the methods based on probabilistic and linear combination, respectively.
We report results on hypotheses collected with
DPM-based methods (\cite{rlopez_2011,GeigerNIPS11}) 
and CNN-based methods (Faster RCNN~\cite{alexnetNIPS12} 
+ viewpoint CNN~\cite{ren15fasterrcnn}).}

\label{table:combinedSingleClassExperiment}
\end{table}


\begin{table}[h!]
\centering

\begin{tabular}{|l|c|c|c|c|c|c}
\cline{1-6}
\multicolumn{6}{|l|}{ \textbf{KITTI dataset \cite{Geiger12KITTI} } - AVP performance } \\
\cline{1-6} 
\multirow{2}{*}{\backslashbox{\scriptsize{LC}}{\scriptsize{RC}} }
		     &  \multirow{2}{*}{None}    & \multicolumn{2}{c|}{Aggressive} & \multicolumn{2}{|c|}{Cautious}  \\ 
		     & & \footnotesize{RF1} & \footnotesize{RF2} & \footnotesize{RF1} & \footnotesize{RF2} \\
\cline{1-6} \cline{1-6}
 None		     	       	 & -    & 0.06 & 0.06 & 0.06 & 0.07 \\
\multirow{2}{*}{Lopez et al.\cite{rlopez_2011} }  & \multirow{2}{*}{0.06} & 0.07 & \textbf{0.08} & 0.08 & 0.08 & \scriptsize{Prob.}\\
						  &  & 0.08 & 0.08 & 0.09 & \textbf{0.11} & \scriptsize{Linear} \\

\hline\hline
None		     		 & -    & 0.10 & 0.14 & 0.09 & 0.10 \\
\multirow{2}{*}{Geiger et al.\cite{GeigerNIPS11}} & \multirow{2}{*}{0.07} & 0.17 & 0.17 & 0.17 & \textbf{0.18} & \scriptsize{Prob.} \\   
				 &   & 0.18 & \textbf{0.20} & 0.19 & 0.19 & \scriptsize{Linear} \\   
				 
\hline\hline
None		     		 & -    & 0.28 & 0.25 & 0.30 & 0.28 \\
\multirow{2}{*}{Ren et al.\cite{ren15fasterrcnn}} & \multirow{2}{*}{0.37} & 0.36 & 0.37 & 0.36 & 0.37 & \scriptsize{Prob.} \\   
				 &   & \textbf{0.39} & \textbf{0.39} & \textbf{0.39} & \textbf{0.39} & \scriptsize{Linear} \\   				 				 
				 
\cline{1-6}     

\end{tabular} 
					   
\caption{ Combined object viewpoint classification performance on  
KITTI \cite{Geiger12KITTI}. 
Average Viewpoint Precision (AVP) performance is presented for the combination of 
the local classifier (LC) and relational classifier (RC) 
using the methods based on probabilistic and linear combination, respectively.
We report results on hypotheses collected with
DPM-based methods (\cite{rlopez_2011,GeigerNIPS11}) 
and CNN-based methods (Faster RCNN~\cite{alexnetNIPS12} 
+ viewpoint CNN~\cite{ren15fasterrcnn}).}

\label{table:combinedSingleClassExperimentAVP}
\end{table}

\textbf{Discussion:}
At first glance, when focusing on the MPPE metric, we can notice that 
the dominance of methods based on cautious inference over those based 
on aggressive inference is not as marked as it was on the purely-contextual 
experiment.
For the case of the combination of local and contextual responses using 
the method based on linear combination (SVM-based), the performance difference between methods
using these two types of inference is reduced to 6 MPPE points. For the case 
when the probabilistic (KDE-based) method is used for combining LC and RC, methods 
based on aggressive inference outperform methods that perform 
cautious inference by 3 MPPE points.
We can verify in Table \ref{table:combinedSingleClassExperiment} 
that the combination of some context-based methods with the local 
classifiers manages to improve the initial performance obtained 
by the local classifier, especially for the case of the DPM-based 
methods (\cite{GeigerNIPS11,rlopez_2011}).
\newtext{
For the case of joint object localization and viewpoint estimation 
(AVP metric), we can notice that superior performance is achieved 
when using the \textit{linear} method for late fusion. Moreover, 
similar as with the MPPE metric, the improvement brought by the 
contextual model is higher over the DPM-based methods ($\sim$10 AVP points) 
than over the CNN-based methods (2 AVP points). This shows that 
the proposed context-based method is not only able to provide 
improvements in terms of viewpoint but also in terms of object 
localization.
}

\newtext{
In Figure~\ref{fig:qualitativeResults} we can notice that by considering 
contextual information (2nd and 3rd columns) we are able to correct some 
of mistakes initially made by the local detector (1st column).
}


\begin{table*}[t]

\centering

\begin{tabu}{|l||c|c|c||c|c|c||c|c|c|}
\hline

 \multicolumn{10}{|l|}{ \textbf{KITTI dataset \cite{Geiger12KITTI}}}  \\ \hline 
 
  \multicolumn{10}{|l|}{ \textbf{Local Information}}  \\ \hline  
 & \multicolumn{3}{|c||}{ \textbf{Lopez et al. \cite{rlopez_2011}}} & \multicolumn{3}{|c||}{ \textbf{Geiger et al. \cite{GeigerNIPS11}}} & \multicolumn{3}{c|}{ \textbf{Ren et al. \cite{ren15fasterrcnn}}}   \\ \hline 
 \textbf{} & \textbf{Low}  &  \textbf{High}  &  \textbf{All} & \textbf{Low}  &  \textbf{High}  &  \textbf{All} & \textbf{Low}  &  \textbf{High}  &  \textbf{All}  \\

 	& 0.07	& 0.05	& 0.06	& 0.07	& 0.07	& 0.07	& 0.39	& 0.36	& 0.37	\\
									
\hline

 \multicolumn{10}{|l|}{ \textbf{Contextual Information}}  \\ \hline  
 & \multicolumn{3}{|c||}{ \textbf{Lopez et al. \cite{rlopez_2011}}} & \multicolumn{3}{|c||}{ \textbf{Geiger et al. \cite{GeigerNIPS11}}} & \multicolumn{3}{c|}{ \textbf{Ren et al. \cite{ren15fasterrcnn}}}   \\ \hline 
 \textbf{Method} & \textbf{Low}  &  \textbf{High}  &  \textbf{All} & \textbf{Low}  &  \textbf{High}  &  \textbf{All} & \textbf{Low}  &  \textbf{High}  &  \textbf{All}  \\  
 	
\hline   
			
Aggressive-RF1	& 0.05	& 0.07	& 0.06	& 0.08	& 0.12	& 0.10	& 0.26	& 0.30	& 0.28	\\
Aggressive-RF2	& 0.04	& 0.08	& 0.06	& 0.10	& 0.16	& 0.14	& 0.22	& 0.28	& 0.25	\\
Cautious-RF1	& 0.05	& 0.07	& 0.06	& 0.06	& 0.10	& 0.09	& 0.30	& 0.31	& 0.30	\\
Cautious-RF2	& 0.05	& 0.08	& 0.07	& 0.08	& 0.12	& 0.10	& 0.27	& 0.29	& 0.28	\\
									
\hline

 \multicolumn{10}{|l|}{ \textbf{Local + Contextual Information (Probabilistic Combination) }}  \\ \hline  
 & \multicolumn{3}{|c||}{ \textbf{Lopez et al. \cite{rlopez_2011}}} & \multicolumn{3}{|c||}{ \textbf{Geiger et al. \cite{GeigerNIPS11}}} & \multicolumn{3}{c|}{ \textbf{Ren et al. \cite{ren15fasterrcnn}}}   \\ \hline 
 \textbf{Method} & \textbf{Low}  &  \textbf{High}  &  \textbf{All} & \textbf{Low}  &  \textbf{High}  &  \textbf{All} & \textbf{Low}  &  \textbf{High}  &  \textbf{All}  \\  
									
Aggressive-RF1	& 0.06	& 0.08	& 0.07	& 0.14	& 0.19	& 0.17	& 0.36	& 0.36	& 0.36	\\
Aggressive-RF2	& 0.07	& 0.09	& 0.08	& 0.15	& 0.19	& 0.17	& 0.38	& 0.37	& 0.37	\\
Cautious-RF1	& 0.07	& 0.09	& 0.08	& 0.14	& 0.20	& 0.17	& 0.36	& 0.35	& 0.36	\\
Cautious-RF2	& 0.06	& 0.10	& 0.08	& 0.15	& 0.20	& 0.18	& 0.38	& 0.36	& 0.37	\\

\hline

\multicolumn{10}{|l|}{ \textbf{Local + Contextual Information (Linear Combination) }}  \\ \hline  
& \multicolumn{3}{|c||}{ \textbf{Lopez et al. \cite{rlopez_2011}}} & \multicolumn{3}{|c||}{ \textbf{Geiger et al. \cite{GeigerNIPS11}}} & \multicolumn{3}{c|}{ \textbf{Ren et al. \cite{ren15fasterrcnn}}}   \\ \hline 
 \textbf{Method} & \textbf{Low}  &  \textbf{High}  &  \textbf{All} & \textbf{Low}  &  \textbf{High}  &  \textbf{All} & \textbf{Low}  &  \textbf{High}  &  \textbf{All}  \\  

Aggressive-RF1	& 0.06	& 0.09	& 0.08	& 0.15	& 0.21	& 0.18	& 0.40	& 0.39	& 0.39	\\
Aggressive-RF2	& 0.06	& 0.09	& 0.08	& 0.17	& 0.22	& 0.20	& 0.40	& 0.39	& 0.39	\\
Cautious-RF1	& 0.08	& 0.11	& 0.09	& 0.16	& 0.22	& 0.19	& 0.40	& 0.39	& 0.39	\\
Cautious-RF2	& 0.10	& 0.12	& 0.11	& 0.16	& 0.22	& 0.19	& 0.40	& 0.39	& 0.39	\\

\hline

\hline 
\end{tabu}

\caption{Average viewpoint precision (AVP) 
for the proposed method for context-based viewpoint 
classification on KITTI \cite{Geiger12KITTI} dataset.
We report results on hypotheses collected with
DPM-based methods (\cite{rlopez_2011,GeigerNIPS11}) 
and CNN-based methods (Faster RCNN~\cite{alexnetNIPS12} 
+ viewpoint CNN~\cite{ren15fasterrcnn}).
Note how the methods focused on local information tend to 
dominate on images with a low number of instances (\textit{Low} subset), 
while the context-based methods have superior performance 
on images with a high number of instances (\textit{High} subset).}

\label{table:objectsPerImageAnalysis}

\end{table*}

\newtext{\subsection{Deeper Analysis}
In order to identify the scenarios in which the proposed contextual approach benefits methods based on local information, we provide an analysis on common viewpoint classification errors (Section \ref{sec:viewpointErrorAnalysis}) and on the effect that the number of occurring object instances has in object viewpoint classification performance (Section \ref{sec:objectsPerImageAnalysis}).
\subsubsection{Viewpoint Classification Error Analysis}
\label{sec:viewpointErrorAnalysis}
We provide now an analysis on common viewpoint classification errors made by both the local and the proposed context-based method.
Following the protocol from \cite{RendondoECCV16} these errors are grouped as "opposite" (viewpoints with a difference of 180 degrees), "nearby" (viewpoints from neighboring viewpoint classes) and "other" (all the other viewpoints). In Figure~\ref{fig:viewpointErrorAnalysis} we show the percentage of instances that belong to each of these three groups plus the percentage of object instances that were predicted correctly. Following the previous experiments, we show the performance when the local classifier (LC), the relational classifier (RC) and the combination of both is employed.
}

\newtext{
\textbf{Discussion:} 
At first glance, Figure~\ref{fig:viewpointErrorAnalysis} further confirms 
the observations made in previous experiments. When focusing on the local 
classifier we can notice that the CNN-based method has a higher number 
of correct predictions (blue) when compared to the DPM-based methods. 
It is also noticeable that the most common type of viewpoint estimation 
error lies in the nearby group (orange) for all the local methods.
}

\newtext{
When focusing on the purely context-based methods (RC), we can notice that 
for all cases the nearby error is reduced. Moreover, we can notice that for 
the case of the DPM-based methods, there is some increase in the percentage 
of object instances classified correctly.
}

\newtext{
Finally, when both local and contextual information is considered, the 
number of correctly predicted instances is further increased for all the 
local methods, especially when using a linear combination method.
For the DPM-based methods the nearby errors are further reduced w.r.t. to results 
obtained in the purely contextual case. Moreover, we notice that for all 
the cases, \textit{other}-type errors get attenuated. 
We can notice that for the case of CNN-based methods, the combination of 
local and contextual information produces a reduction of the 
\textit{nearby}-type errors while keeping the \textit{opposite} and 
\textit{other} errors at the level of the local method. 
}

\newtext{
The previous observations in combination with Figure~\ref{fig:viewpointErrorAnalysis} 
provide insight on the benefits that the proposed contextual method brings 
to the methods based on local information.  First, for the case of DPM-based methods 
(\cite{GeigerNIPS11, rlopez_2011}), the context-based methods bring improvement in performance by 
reducing errors of type \textit{nearby} and \textit{other}. Moreover, for 
the case of \cite{GeigerNIPS11}, the proposed context-based method is able 
to also address errors between opposite viewpoints.
Finally, for the case of the CNN-based method, contextual information complements 
methods based on local information by reducing errors between nearby viewpoints.
}

\newtext{
\subsubsection{Effect of the Number of Objects per Image}
\label{sec:objectsPerImageAnalysis}
The proposed method fully relies on the exploitation of relations between objects (Section \ref{sec:representation}). Since these relations are directly defined from object instances occurring (or detected) in images, in this experiment we measure the effect that the number of object instances has on the effectiveness of context-based methods.
Towards this goal, in this experiment we measure performance in two independent subsets of images. In order to define these subsets, we first compute the average number of annotated object instances per image in the dataset. In the case of the KITTI dataset this average value is 5 objects per image. The \textit{Low} subset, is composed by images with a number of annotated object instances less or equal to the average, while the \textit{High} subset is composed by the images with higher number of objects. We report the AVP performance of these two subsets and the subset composed by \textit{All} the images in Table \ref{table:objectsPerImageAnalysis}.
}

\newtext{
\textbf{Discussion:}
We can notice that when only considering local information, higher performance is achieved when focusing on images with few object instances, which likely implies lower inter-object occlusion and a lower number of instances with small size. On the contrary, when only considering contextual information, higher performance is achieved always for the \textit{High} subset. 
This suggests that while the local classifier is able to handle effectively images with few objects (possibly with large size and low level of occlusions), the contextual information is able to compensate for some of the hard scenarios (low object size and occluded objects) of the local classifier.
Finally, for the case when local and contextual information is combined, we can notice that for the DPM-based local methods, the performance on the \textit{High} subset is still superior to that on the \textit{Low} subset. For the CNN-based local method we notice that the performance in both \textit{Low} and \textit{High} subsets tend to be comparable. This is to be expected since the DPM-based methods are less effective than the CNN-based methods, hence leaving more room for improvement for the contextual model. For the linear combination with the CNN-based method, we notice that the contextual model is able to push performance of the local classifier on the \textit{High} subset for  $\sim$3 AVP points while only 1 AVP point of the case of the \textit{Low} subset. This shows that the proposed method has the potential of improving both object localization and viewpoint estimation performance especially for the case of crowded scenarios, i.e. images with high number of instances, where inter-object occlusions and instances with small size are likely to occur.
}

\newtext{
\section{Limitations and Future Work}
\label{sec:limitationsFutureWors}
We have presented a method to perform object viewpoint 
estimation by using contextual information derived from the other 
object instances occurring in the scene. Our empirical results show 
that there is a clear potential for this type of contextual cue on 
improving object viewpoint classification. Moreover, we have shown 
that considering contextual information reduces prediction errors 
of type \textit{nearby} and \textit{other}  (Section~\ref{sec:viewpointErrorAnalysis}).
However, in its current state, the proposed method has some limitations 
that could be addressed in further work. These limitations are closely 
related to the two main assumptions behind our method.
First, our model assumes there is a structured behavior (occurrence, viewpoint, localization, etc.) between entities (objects) in the scene. Second, during test time it is required for the local detector to have an acceptable level of performance in order to feed our contextual model with informative cues and to be able to be improved from. 
}

\newtext{
Regarding the first assumption, in our experiments, the KITTI dataset resembles a Urban setting and due its large volume of data, multiple plausible scene states (object occurrence, location, size, viewpoints, etc.) are presented. Moreover, this is done for different scene types, i.e. highway, urban, residential, etc. Hence, the KITTI dataset constitutes a representative dataset for the setting to be modeled. 
On the contrary, for the case of a more biased dataset (either at object locations, viewpoints, etc.), there might not be sufficient representative information to accurately model the relational behavior of the objects in the scene.
This is in line with the findings from the Collective 
Classification and Link-based Classification communities \cite{bilgicICDM07,JensenLinkDensity03,NevilleLGM2005,Senaimag08}. 
Those findings state that methods that reason about links, 
or relations, between entities have an improved performance 
when operating on a setting with high link density.
It is important to notice that link density does not only refer 
to the quantity (amount) of the links between entities, 
but also refers to the quality (representativeness) of such links.
In our particular setting, link density is directly influenced 
by the objects occurring in the images, more specifically, the 
relations between them, i.e. the relative locations and viewpoints 
in which these objects co-occur. 
In this regard, the KITTI dataset has a high amount of links while 
being diverse and covering different scenarios of co-occurring object 
viewpoints.
Based on this, a reasonable hypothesis is that the proposed context-based method 
may underperform when learning object relations from biased datasets showing 
poor link density. 
}

\newtext{
Regarding the second assumption, when using a low-performing local detector, 
while allowing significant room for improvement, it will introduce noise to 
the contextual model and further complicate the following processes. On the contrary, 
having a highly-performing detector will significantly reduce the cases in which 
the context model could produce improvement. In our experiments we have noted 
that by going from DPM-based viewpoint-aware detectors to CNN-based methods we 
can increase the overall performance while still being able to bring improvement 
via the proposed context-based method.
This assumption is related to the ratio between true and false positives 
predicted by the local classifier.
\newtextRed{This ratio is important if we consider that the number of 
relations has a quadratic growth w.r.t. the number of object 
hypotheses}, hence introducing a significant amount of noise in 
the context-based classification process.
This ratio is known as \textit{class skewness}, or 
\textit{labeled proportion}, in the collective classification literature \cite{ChakrabartiSkewness1998,MacskassyP07,McDowellCautiousGA09,Taskar2002}, 
more specifically when focusing on within network 
classification tasks where predictions about some 
nodes are based on other nodes. In this type of 
tasks, class skewness measures the proportion of data 
that is known, or predicted with certainty, w.r.t. the 
whole data. In scenarios where class skewness is low, 
there is not enough certain information to guide the 
inference process.
In scenarios with high class skewness, the 
performance of collective classification is better or 
comparable to that of local classification.
}

\newtext{
An additional weak point of the proposed method lies in the way in which it combines 
the responses from the local and relational classifiers.
A more structured approach to improve the combination of these responses is by formulating 
the object viewpoint classification problem within a Conditional Random Field (CRF) setting. 
This structured setting could be defined over an undirected graph $G=( V , E )$ where the nodes $V$ are the objects and edges $E$ are the relations between objects (as defined in Section \ref{sec:pairwiseRelations}). In this case the states that a node (object) may take are defined by the possible viewpoint values $\alpha$.
The Unary potential is defined by the local evidence given by the object detector -- more specifically, the confidence of the classifier over different object viewpoints. The Pairwise potential of the edges is defined by the distribution of a relation $r_{ij}$ occurring between objects $o_i$ and $o_j$ over different viewpoint values. This potential is estimated in a similar fashion as the pairwise term $v( o_i , o_j )$ from Equation~\ref{eq:relationProb}.
Having these potentials in place, we could solve the CRF in order to find the global optimal configuration that satisfies both the local response given by the object detector as well as the object relations considered by our model.\\ 
On the positive side, following this setting there are several factors that can be evaluated, e.g. graph topology, regularization, edge representation, that can be optimized in order to push performance further. However, on the weak side,  the proposed CRF-method has the disadvantage of being more computationally expensive when compared to the proposed method.\\  
}

\newtext{
Not withstanding these limitations and regardless of the collective classification method that is used, we have shown that performing collective inference  results in performance gains on object viewpoint estimation. 
%
This finding becomes even more relevant when we keep in mind the fact that the proposed context-based method has proven to have complementarity w.r.t. local methods, and that all this has been achieved while starting from one of the simplest collective classification methods. This suggests that reasoning about object relations in 2D image space can indeed assist the task of object viewpoint classification, and that there is a clear potential for more advanced methods for collective classification to further improve object viewpoint estimation performance.
%
}

\section{Conclusions}
\label{sec:conclusions}

In this paper we have presented a method to
exploit contextual information, in the form 
of relations between objects, for object viewpoint 
classification in the 2D image space. 
Our experiments show that even when contextual 
information alone cannot solve the viewpoint 
estimation problem accurately, it is able to 
provide good priors about the location and viewpoint 
of objects. 
This characteristic can be useful for tasks 
such as object proposal generation driven 
by object relations. 
In addition, our experiments show that in the 
absence of local appearance information about 
the object to be classified, performing cautious 
inference about object relations outperforms 
its aggressive counterpart. 
Our analysis reveals that the proposed 
context-based method complements methods 
that only consider local information in two ways:
First, it is able to reduce viewpoint errors related 
to nearby and other non-opposite types. 
Second, it produces superior performance in 
settings where a high number of object instances 
occur.
Investigating more structured methods to enforce 
global consistency and to model object relations 
between multiple object categories constitutes 
our next steps for future work.

\vspace{4mm}
\noindent\textbf{Acknowledgments:}
This work was supported by the FWO project SfS, 
the KU Leuven PDM Grant PDM/16/131, 
and a NVIDIA Academic Hardware Grant.






\bibliographystyle{elsart-num-sort}
\bibliography{egbib}








\end{document}